\documentclass[11pt,a4paper]{ouparticle}

\usepackage{hyperref}
\newcommand\figsubref[2]{\hyperref[#1]{\ref*{#1}#2}}

\hypersetup{
  linkcolor  = BrickRed,
  citecolor  = NavyBlue,
  urlcolor   = Aquamarine,
  colorlinks = true,
}

\let\includesvg\includegraphics

\begin{document}

\title{A Brain-like Synergistic Core in LLMs\\ Drives Behaviour and Learning}

\author{
    Pedro Urbina-Rodriguez$^{1,2, *}$, Zafeirios Fountas$^{2}$, Fernando E. Rosas$^{4,5,6}$, Jun Wang$^{3}$, Andrea I. Luppi$^{6,7,8}$, Haitham Bou-Ammar$^{2,3}$, Murray Shanahan$^{1}$, \mbox{Pedro A. M. Mediano$^{1,9}$}
\\[1ex]
\small
\\[1ex]
\small
    $^{1}$Department of Computing, Imperial College London, London, UK;
    $^{2}$Huawei, Noah's Ark Lab, London, UK;
    $^{3}$AI Centre, Department of Computer Science, University College London, London, UK;
    $^{4}$Department of Informatics, University of Sussex, Brighton, UK;
    $^{5}$Centre for Complexity Science and Center for Psychedelic Research, Department of Brain Science, Imperial College London, London, UK;
    $^{6}$Department of Psychiatry and Centre for Eudaimonia and Human Flourishing, University of Oxford, Oxford, UK;
    $^{7}$Division of Information Engineering and St John's College, University of Cambridge, Cambridge, UK;
    $^{8}$Montreal Neurological Institute, McGill University, Montreal, CA;
    $^{9}$Division of Psychology and Language Sciences, University College London, London, UK
    \\[1ex]
    * Correspondence: \texttt{urbina2000pedro@gmail.com}
}


\abstract{
    The independent evolution of intelligence in biological and artificial systems offers a unique opportunity to identify its fundamental computational principles. Here we show that large language models spontaneously develop synergistic cores -- components where information integration exceeds individual parts -- remarkably similar to those in the human brain. Using principles of information decomposition across multiple LLM model families and architectures, we find that areas in middle layers exhibit synergistic processing while early and late layers rely on redundancy, mirroring the informational organisation in biological brains. This organisation emerges through learning and is absent in randomly initialised networks. Crucially, ablating synergistic components causes disproportionate behavioural changes and performance loss, aligning with theoretical predictions about the fragility of synergy. Moreover, fine-tuning synergistic regions through reinforcement learning yields significantly greater performance gains than training redundant components, yet supervised fine-tuning shows no such advantage. This convergence suggests that synergistic information processing is a fundamental property of intelligence, providing targets for principled model design and testable predictions for biological intelligence.
}

\date{}


\maketitle


Understanding how intelligent systems integrate information to produce capabilities beyond their constituent parts remains a fundamental challenge in both neuroscience and artificial intelligence. From populations of neurons giving rise to consciousness, to linguistic representations that capture abstract semantic relationships, intelligence appears linked to the integration of information. Yet we have historically lacked rigorous tools to distinguish genuine integration -- whereby the whole becomes more than the sum of its individual parts -- from mere aggregation of independent signals. Recent advances in information theory now provide such a framework: the information-theoretic construct of synergy -- information available only through joint observation of multiple variables -- precisely quantifies whether a system synthesises inputs into integrated representations~\cite{pid2010beer,gawne1993}, offering a principled lens for analysing the computational architecture of intelligence.

\begin{figure}[!ht]
    \centering
    \includegraphics[width=0.9\linewidth]{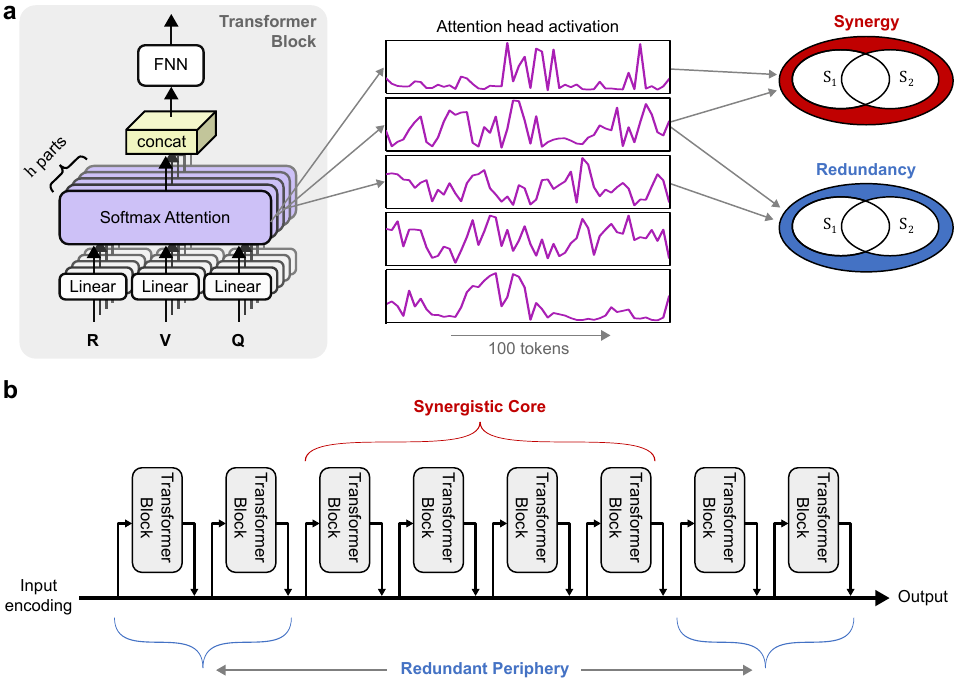}
    \caption{\small\textbf{Schematic diagram of the paper.} \textbf{(a)} Methods: We measure the attention head activation of each attention heads in each transformer block, during the generation of a sequence of 100 tokens. Pairs of activation time series are considered as sources of information ($S_1$, $S_2$) and used to calculate the information-theoretic measures of synergy and redundancy. \textbf{(b)} Key result: LLMs contain a ``synergistic core'' in their middle layers, while their early and late layers (closer to input and output, respectively) are predominantly redundant. Through observational and interventional experiments, we demonstrate the relevance of the synergistic core for model behaviour and generalization.}
    \label{fig:conceptual}
\end{figure}

Building on Shannon's foundational information theory~\cite{shannon_1948}, Partial Information Decomposition (PID) formalises this by distinguishing three fundamental information types: redundant (accessible through multiple sources independently), unique (specific to individual sources), and synergistic (emerging only from joint observation)~\cite{pid2010beer}. Stereoscopic depth perception exemplifies synergy: in humans, neither eye alone encodes three-dimensional structure, yet their integration creates genuinely new spatial information. Synergistic processing appears crucial for flexible intelligence, enabling systems to integrate information from multiple modalities and learn diverse tasks~\cite{proca2024}, with synergistic dynamics tightly coupled to predictive performance~\cite{tolle2024evolving}. This framework enables quantitative comparison of information processing across biological and artificial systems \cite{Tax2017,ehrlich2023complexity}.

In the human brain, PID analysis has identified a so-called \textit{synergistic core} encompassing the default mode and executive control networks~\cite{luppi2022synergistic} -- regions associated with complex cognition and a `global workspace' where different information streams converge, which are situated at the top of the cortical processing hierarchy and have undergone the greatest degree of evolutionary expansion between humans and other primates \cite{baars1993,dehaene2001workspace,luppi2024workspace,luppi2022synergistic,wei2019genetic,varley2023partial}. This core exhibits high global efficiency for rapid integration across distributed regions, and elevated synergistic interactions compared to sensorimotor (i.e., input and output) areas. Synergistic processing shows evolutionary increases, with humans demonstrating significantly higher synergy than non-human primates~\cite{luppi2022synergistic}. These findings establish synergistic organisation as a hallmark of sophisticated human cognition that is causally linked to behavioural capabilities. This raises a testable question: do artificial intelligence systems that achieve complex capabilities develop similar information-theoretic organisation, and does this contribute functionally to their performance?

Here we answer in the affirmative, and demonstrate that today's most advanced AI systems -- large language models (LLMs) -- spontaneously develop brain-like synergistic cores through learning. Applying Integrated Information Decomposition ($\Phi$ID)~\cite{phid2021} -- an extension of PID to temporal dynamics -- we treat attention heads and mixture-of-experts (MoE) modules as information-processing units and analyse their interactions across diverse cognitive tasks. Middle layers consistently exhibit predominantly synergistic interactions, while early and late layers rely on redundant information processing -- reminiscent of the redundancy-dominated sensory and motor regions of the human brain. This pattern emerges robustly across model families (Gemma~\cite{google2024gemma}, Llama~\cite{dubey2024llama}, Qwen \cite{yang2025qwen3}, DeepSeek \cite{liu2024deepseek}) and fundamentally different architectures (transformer attention and mixture-of-experts), yet is absent in randomly initialised networks, confirming that learning drives synergistic structure development. The spatial consistency across architectures -- and convergence with biological intelligence -- suggests a computational necessity rather than an architectural artefact.

To foreshadow our results, we show that this synergistic organisation is functionally relevant through two complementary approaches. First, systematic ablations reveal that removing high-synergy components causes disproportionate performance degradation compared to removing redundant or random components with identical parameter counts. Second, and more compellingly, targeted reinforcement learning (RL) of synergistic components yields significantly greater performance improvements than training redundant regions. Interestingly, supervised fine-tuning shows no such difference, suggesting a selective role of synergy in learning paradigms that promote generalisation (RL) but not memorisation (supervised)~\cite{chu2025sft}. Graph-theoretic analysis reveals that synergistic cores exhibit high global efficiency characteristic of integrative processing, while redundant cores show modular organisation typical of specialised processing -- recapitulating biological network architecture~\cite{luppi2022synergistic}.

The independent emergence of synergistic information processing in biological and artificial intelligence -- despite vastly different substrates and learning rules -- suggests a fundamental computational principle. For AI development, identifying synergistic cores offers paths toward more efficient training through targeted parameter updates, principled model compression by preserving high-synergy components, and enhanced interpretability by linking function to information dynamics. For neuroscience, parallel principles in artificial systems provide computational validation for theories of brain organisation and generate testable predictions: synergistic circuits should show greater plasticity during reinforcement learning than supervised learning, and flexible task transfer should depend more critically on synergistic than redundant regions. In what follows, we establish the existence and properties of synergistic cores in LLMs, demonstrate their functional necessity through ablation and learning experiments, and explore how these findings illuminate computational principles underlying intelligence.

\section*{Results}\label{sec:results}

In order to explore the informational architecture of LLMs, we closely adopt the conceptual and methodological tools used by Luppi et al. \cite{luppi2022synergistic}. We consider LLMs as distributed information-processing systems, where either attention heads or experts (in mixture of experts [MoE] architectures) are the relevant sub-units of the system. This approach allows us to apply the recently developed Integrated Information Decomposition ($\Phi$ID) framework \cite{phid2021} to study information flow between attention heads through time. 
Following Luppi et al. \cite{luppi2022synergistic}, we use $\Phi$ID to investigate the properties of temporally persistent synergy and redundancy as our notions of synergy and redundancy, respectively.

\subsection*{LLMs Develop a Synergistic Information Processing Core}

The first question we seek to answer is how synergy and redundancy are distributed across LLMs. 
To quantify the synergy and redundancy of LLMs, we record attention head activations while models process diverse prompts spanning 6 cognitive task categories (detailed in Appendix \ref{appendix:prompts}). For each prompt, we generate a 100-token response and record the L2 norm of attention outputs for each head across all response tokens. Similarly, for LLMs with MoE architecture, we compute the L2 of the expert outputs. This procedure yields time series data for each attention head or expert on each prompt (Fig.~\figsubref{fig:conceptual}{a}).

\begin{figure}[!ht]
    \centering
    \includegraphics[width=0.8\linewidth]{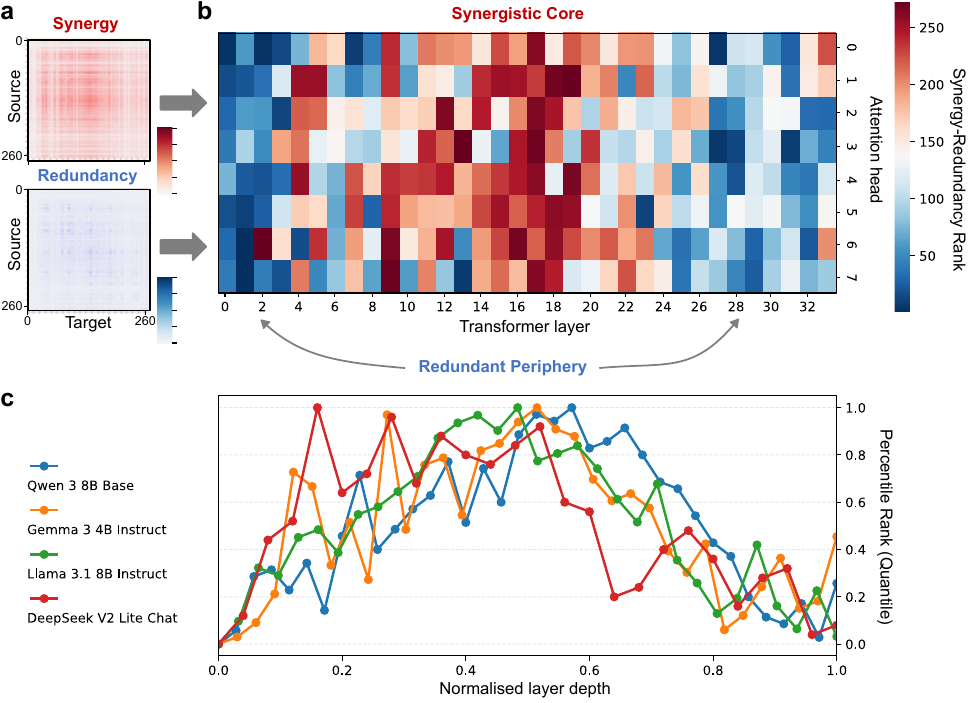}
    \caption{\small\textbf{LLMs contain a synergistic core comprising their middle layers}. \textbf{(a)} Heatmaps showing synergy and redundancy between pairs of attention heads in Gemma 3 4B. \textbf{(b)} Heatmap showing the synergy-redundancy ranks in Gemma 3 4B. High (red) values indicate that an attention head's interactions are predominantly synergistic, while low (blue) values indicate they are primarily redundant. \textbf{(c)} Synergistic cores across several LLMs, with the y-axis representing the synergy-redundancy rank. Layer numbers are normalised from $0$ to $1$, and the synergy-redundancy rank is normalised using min-max scaling to enable comparisons across models of different sizes. The middle layers of the models exhibit the most synergistic interactions. DeepSeek V2 Lite's synergistic core is computed at an expert level, whereas the rest of models are based on attention heads.}
    \label{fig:synergistic_core}
\end{figure}

With per-attention-head activation time series in hand, we compute the average synergy and redundancy across time and prompts of all pairs of attention heads using the implementation provided by Mediano et al. \cite{phid2021, phid_repo}. Following Luppi et al. \cite{luppi2022synergistic}, we then average the synergy and redundancy values across all attention head pairs involving a given individual head, yielding an estimate of how synergistic or redundant each head's interactions are overall. By ranking attention heads based on their synergy and redundancy levels, and then subtracting these ranks, we obtain a \textit{synergy-redundancy rank} that orders attention heads across the whole system from most synergistic to most redundant.

By calculating the average synergy-redundancy rank per layer, results reveal that middle layers of LLMs are dominated by synergistic information processing, while early and late layers
are predominantly characterized by redundant interactions (Fig.~\figsubref{fig:synergistic_core}{b}). This finding is observed consistently across multiple model families and architectures. In particular, for the MoE DeepSeek V2 Lite, the synergistic core is computed using experts instead of attention heads and the same pattern emerges (Fig.~\figsubref{fig:synergistic_core}{c}).

We term this phenomenon a \textit{synergistic core}, following the framework of Luppi et al. \cite{luppi2022synergistic}, and refer to the remaining early and late layers as the \textit{redundant periphery}. 
These results align with established findings in the interpretability community that early layers perform detokenisation, late layers perform tokenisation, while middle layers handle high-level computation \cite{elhage2022solu, nanda2023factfinding}. Additionally, middle layers have been found to contain the largest concentration of multilingual features, a pattern that strengthens with increasing model size \cite{lindsey2025biology}.

\subsection*{Synergistic Core Emerges through Training}

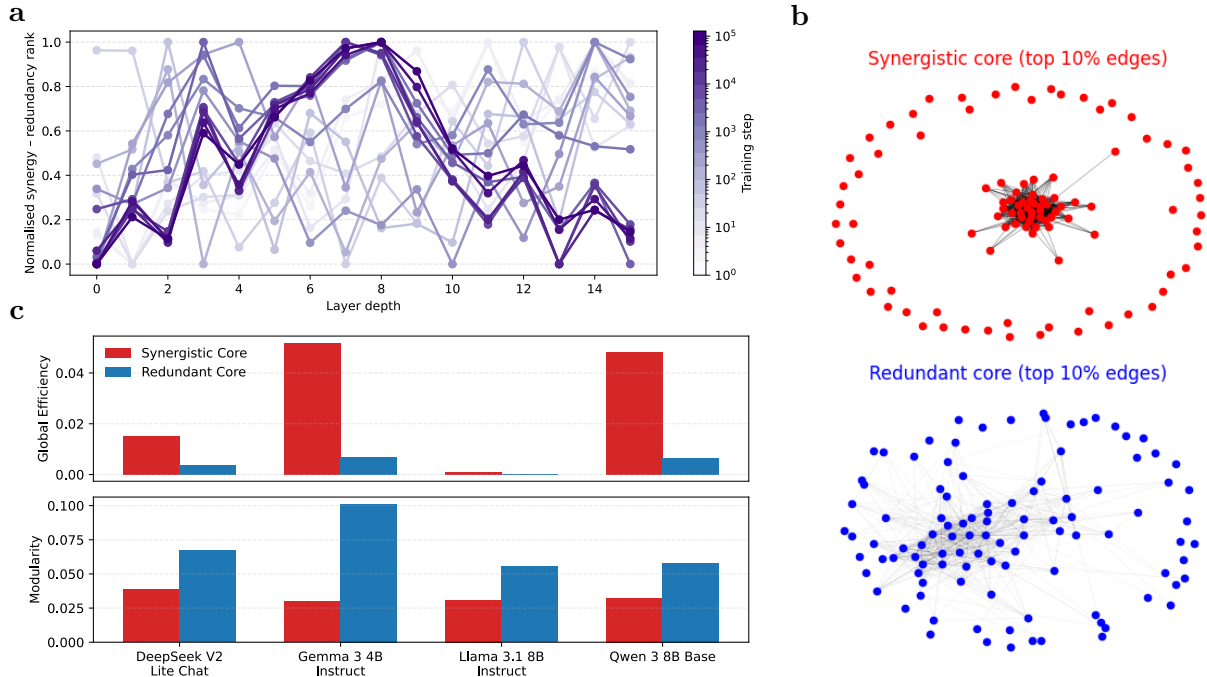
\begin{figure}[!tbp]
    \centering

    \begin{minipage}{0.61\linewidth}
        \includesvg[width=\linewidth]{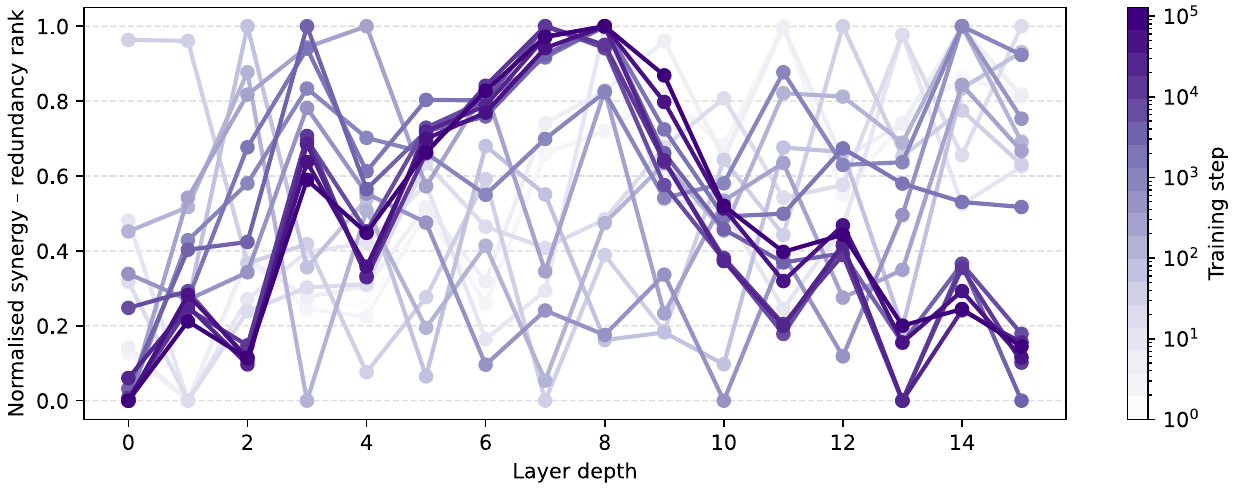}
        \vspace{-1em} 

        \includesvg[width=\linewidth]{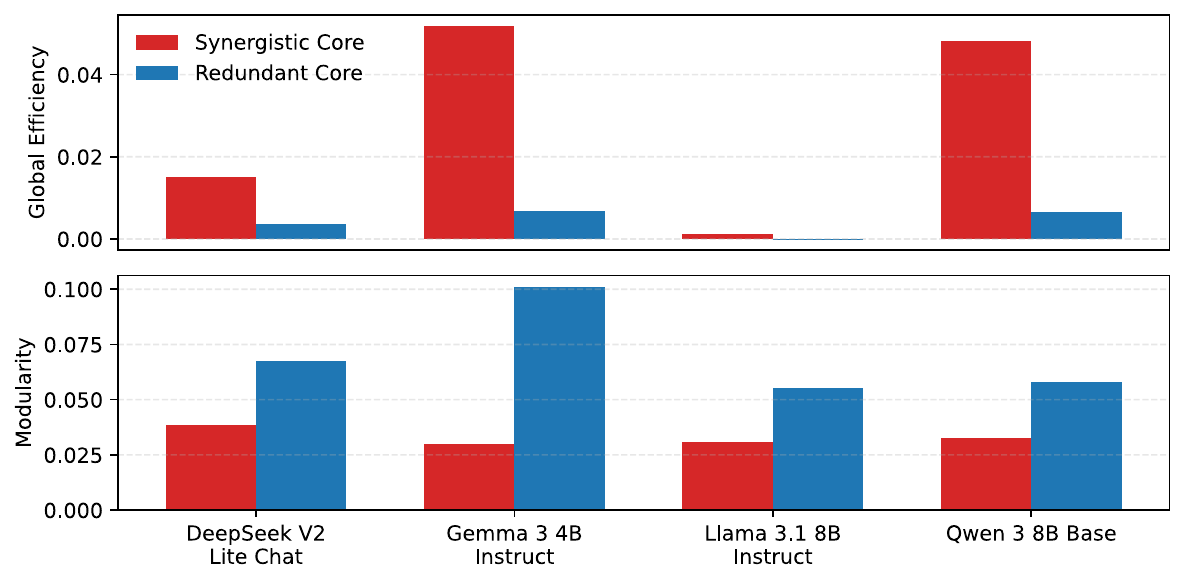}
    \end{minipage}%
    \hfill
    \begin{minipage}{0.35\linewidth}
        \centering
        \includegraphics[width=\linewidth]{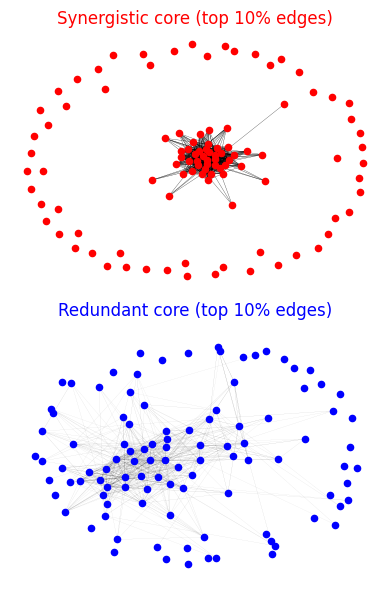}
    \end{minipage}

    \begin{tikzpicture}[overlay]
        \node at (-8, 8.95) {\textbf{a}};
        \node at (-8, 5) {\textbf{c}};
        \node at (2.3, 8.95) {\textbf{b}};
    \end{tikzpicture}

    \caption{\small
        \textbf{LLM's synergistic core emerges through training and shares topological features with the human brain}.
        \textbf{(a)} Normalised synergy–redundancy rank through training for Pythia-1B~\cite{biderman2023pythia}. 
        \textbf{(b)} Synergistic and redundant cores for Gemma 3 1B Instruct \cite{gemma_2025} represented as undirected graphs. Only the top 10\% strongest connections are shown for clarity. 
        \textbf{(c)} Graph-theoretical properties of the synergy and redundancy matrices for different LLMs. 
        For DeepSeek V2 Lite, the matrices are computed expert-wise, and for the remaining models they are computed attention-head-wise. 
    }
    \label{fig:graph}
\end{figure}

After identifying this synergistic core in LLMs, we next investigate its origins. One possibility is that this pattern reflects generic information-theoretic properties of the transformer architecture~\cite{vaswani2017attention}, largely independent of learning or task performance, rather than a signature of learned information processing associated with intelligent behavior.

To distinguish between these alternatives, we analyze how the informational architecture evolves throughout training. Crucially, while the transformer architecture remains fixed, the model’s functional organization and performance change substantially over training. If the synergistic core is primarily an architectural artifact, we would expect it to be present—and largely stable—across training checkpoints, including at early stages prior to meaningful learning. Conversely, if it reflects learned information integration, it should emerge or strengthen progressively during training, tracking improvements in model performance.

To test these predictions, we study the Pythia-1B model~\cite{biderman2023pythia}, which provides model snapshots at multiple training checkpoints, allowing us to explicitly incorporate the temporal (training) dimension into our information-theoretic analysis.

The results are shown in Figure~\figsubref{fig:graph}{a}, analogous to Figure~\figsubref{fig:synergistic_core}{c} but with color indicating training progression. At the earliest training steps, the inverted-U pattern is absent, and the information-theoretic structure appears random. As training progresses, the inverted-U shape gradually emerges and stabilizes, culminating in the fully formed pattern of the final trained model.

These results show that the synergistic core is not an intrinsic property of the transformer architecture itself, but instead emerges through the learning process. This suggests that the synergistic core may be a marker of learned competence or intelligence, a perspective we explore in later sections.

\subsection*{Topological Properties of LLMs' Synergistic Core Match Biological Brains}

After confirming that the synergistic core is associated with intelligent behaviour, we investigate how synergy and redundancy are structured across attention heads. 
To do this, we use the synergy and a redundancy values per pair of attention heads (or experts) as the weights of undirected weighted networks (Fig.~\figsubref{fig:graph}{b}). This allows us to study the topological properties of the synergistic core and the redundant periphery.

Results show that the synergistic core exhibits greater global efficiency consistently across all examined models, whereas the redundant core consistently demonstrates higher modularity (Fig.~\figsubref{fig:graph}{c}). These distinct 
signatures match the functional properties of synergy and redundancy: high global efficiency enables the synergistic core to achieve effective information integration through inter-node communication, while higher modularity reflects strong cohesion among nodes within the redundant periphery.

Remarkably, these observations align precisely with patterns identified by Luppi et al. \cite{luppi2022synergistic} in the human brain, with a highly efficient synergistic network and a modular network of redundant interactions -- suggesting a compelling parallel between artificial and biological intelligence systems.

\subsection*{The Synergistic Core is Functionally Critical}

\begin{figure}
    \centering
    \includesvg[width=0.5\linewidth]{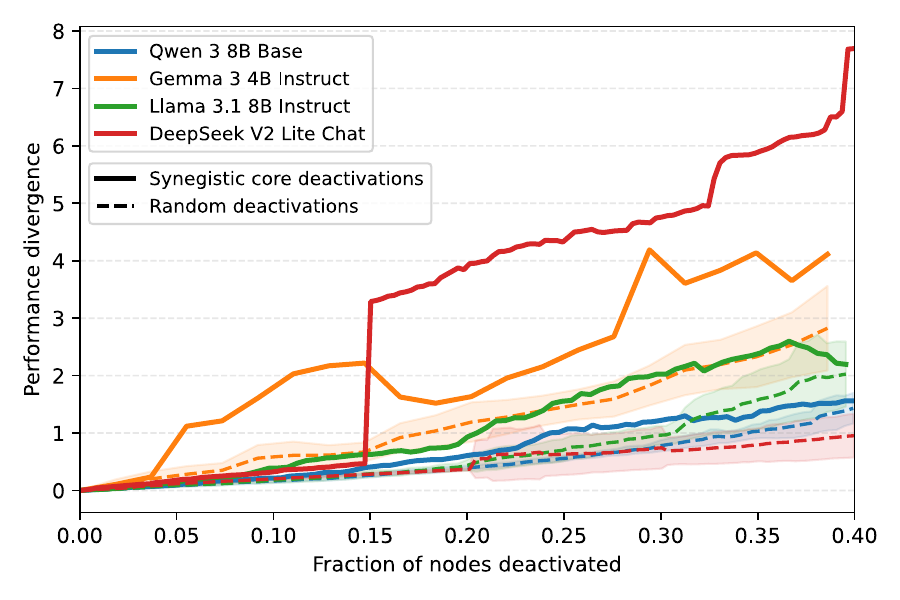}
    \includesvg[width=0.45\linewidth]{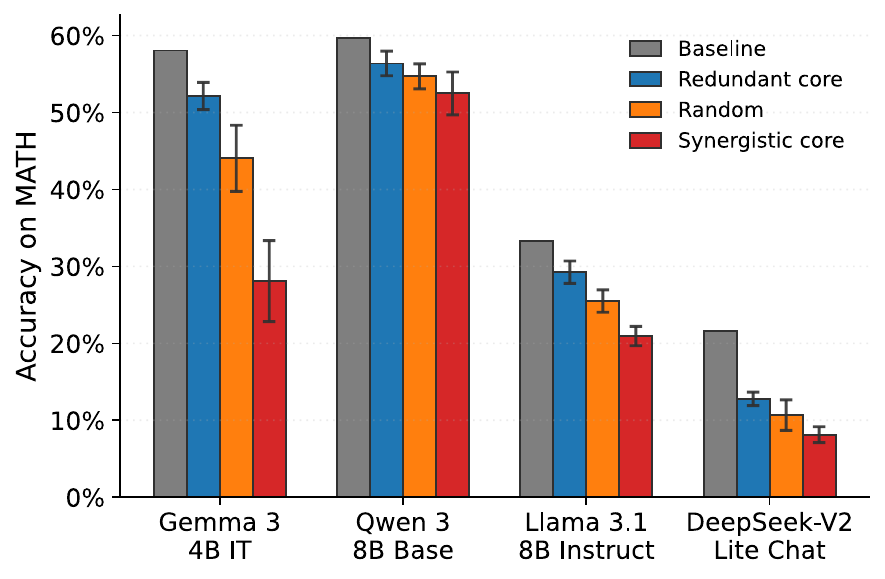}
    
    \begin{tikzpicture}[overlay]
        \node at (-8, 5.5) {\textbf{a}};
        \node at (0.6, 5.5) {\textbf{b}};
    \end{tikzpicture}
    \vspace{-1.5em}
    \caption{\small
        \textbf{Ablating the synergistic core causes greater changes in behaviour and drops in performance.}
        \textbf{(a)} Behaviour divergence as a function of the fraction of nodes deactivated (experts for DeepSeek V2 Lite and attention heads for the remaining models), quantified as the KL divergence between the token strings of original and ablated models. Solid curves correspond to ablating nodes in synergistic order, while dashed curves correspond to randomly ordered ablating. The shaded regions around the dashed curves indicate the standard deviation across five random-order runs.
        \textbf{(b)} Comparison of accuracy on the MATH benchmark \cite{hendrycks2021measuringmathematicalproblemsolving} when perturbing the synergistic core, the redundant core, or random subsets of each model. Perturbations were applied by injecting Gaussian noise into the query/output projections of selected attention heads, or into full expert parameters for MoE architectures.
    }
    \label{fig:deact}
\end{figure}
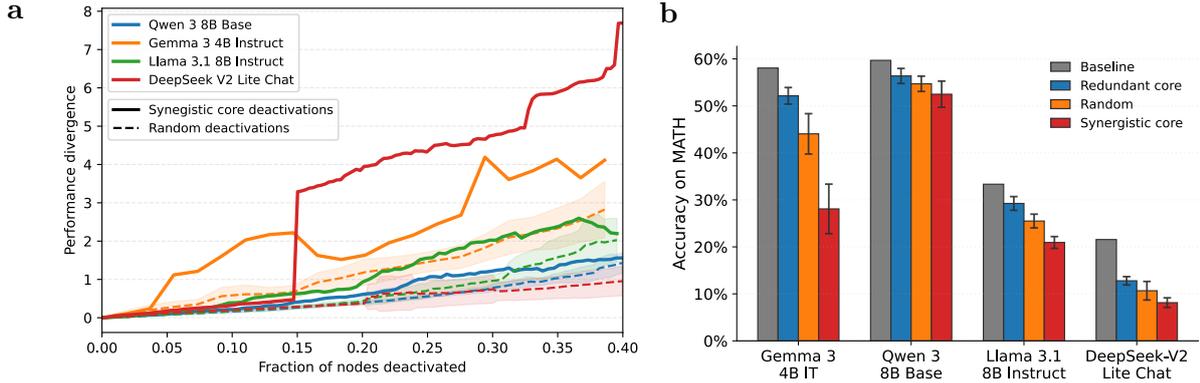

\subsubsection*{Ablating the Synergistic Core Causes Catastrophic Change in Behaviour}

So far, all our results have been purely observational: descriptions of the structure of the synergistic core as the model responds to prompts.
To go beyond observational analyses and understand the functional importance of the synergistic core, we perform two complementary analyses. 

First, we causally perturbed different attention heads and measured their contribution to task behaviour through the Kullback-Leibler (KL) divergence between the resulting distributions of the non-ablated and the ablated model as a proxy to model performance (see \nameref{sec:behaviour-divergence-methods} for details). Our results show that ablating the heads with highest synergy-redundancy rank has a greater impact in performance than ablating them in random order (Fig.~\figsubref{fig:deact}{a}). These results are consistent along the four different open source models, both at an expert level for DeepSeek V2 Lite Chat and at an attention head level for the rest of models. 

To further test its functional relevance, we evaluate how disrupting the synergistic core affects downstream task performance using the \texttt{MATH} dataset \cite{hendrycks2021measuringmathematicalproblemsolving}.
In this experiment, we compared the effect of perturbing the synergistic core, the redundant core, or random subsets of the model by
injecting Gaussian noise into the selected components. For attention heads, noise is applied to the query-projection rows and output-projection columns associated with each head; for MoE experts, it was applied to all parameters within the targeted expert modules. 
 
Results show that perturbing the synergistic core (i.e. the top 25\% most synergistic attention heads according to the synergy–redundancy ranking) leads to a substantially larger degradation in performance than perturbing either random heads or the redundant core (see Figure~\figsubref{fig:deact}{b}). In contrast, perturbing the redundant core has the smallest effect. These findings further support our claim that synergy highlights computationally important components of the model and may serve as an indicator of higher-order computation and cognition within LLMs. Conversely, it stands to reason that lesioning redundant elements should have little effect, since this is precisely the computational role of redundancy: providing robustness.

\subsubsection*{Synergy Enhances Generalisation in Fine-Tuning}

Finally, we seek to understand whether synergistic cores can be causally modified not only to degrade, but to enhance a model's performance.
To do this, we investigate how targeting the synergistic core during reinforcement learning fine-tuning (RLFT) might influence model performance.\footnote{Due to computational limitations, our experiments focus on fine-tuning rather than full-scale pretraining. Nevertheless, we hypothesise that similar patterns and insights might emerge during pretraining, a promising direction for future research.} Interestingly, Luppi et al. \cite{luppi2022synergistic} find that the human brain's synergistic core comprises regions that predominantly activate during higher-order cognitive tasks. Thus, as an analogy, we hypothesise that selectively fine-tuning the synergistic core of LLMs could result in more significant performance improvements in complex cognitive tasks.

Motivated by this neuroscientific analogy, we test our hypothesis by fine-tuning base models under two post-training regimes: supervised fine-tuning (SFT), which is often associated with memorisation, and reinforcement learning fine-tuning (RLFT), which has been shown to better support generalisation \cite{chu2025sft}. Within both regimes, we compared `synergistic fine-tuning,' in which only the top $50\%$ most synergistic heads are updated, against `redundant fine-tuning,' in which only the $50\%$ most redundant heads are updated. These two approaches are compared against randomly updating $50\%$ of the heads irrespectively of their rank. 

For investigating the SFT setting, we use the OpenMathInstruct-2 dataset \cite{toshniwal2024openmathinstruct2}. 
Our results show no significant differences among the three training methods in this setting. We attribute this lack of differentiation to recent findings indicating that foundation models tend to memorise rather than generalise under SFT \cite{chu2025sft}. In this regime, LLMs may store memorised facts in any region of their parameter space, irrespective of distinctions between the synergistic and redundant cores. 

Motivated by findings that show that RLFT is more effective at promoting generalisation~\cite{chu2025sft}, we investigate the RL setting using R1-style post-training \cite{deepseekai2025deepseekr1incentivizingreasoningcapability} using GRPO \cite{shao2024deepseekmathpushinglimitsmathematical} on the MATH train set \cite{hendrycks2021measuringmathematicalproblemsolving} (see \nameref{sec:rl-procedure} for more details). We selected Qwen2.5-Math-1.5B \cite{yang2024qwen25mathtechnicalreportmathematical} as our base LLM for RLFT due to its strong mathematical performance and computational efficiency enabled by its non-Chain-of-Thought (non-CoT) response format. 
Under this setup, our results show that training the synergistic core results in significantly higher performance than training only the redundant core or a random subset of the model.

\begin{figure}[t!]
    \centering
    \includegraphics[width=0.9\textwidth]{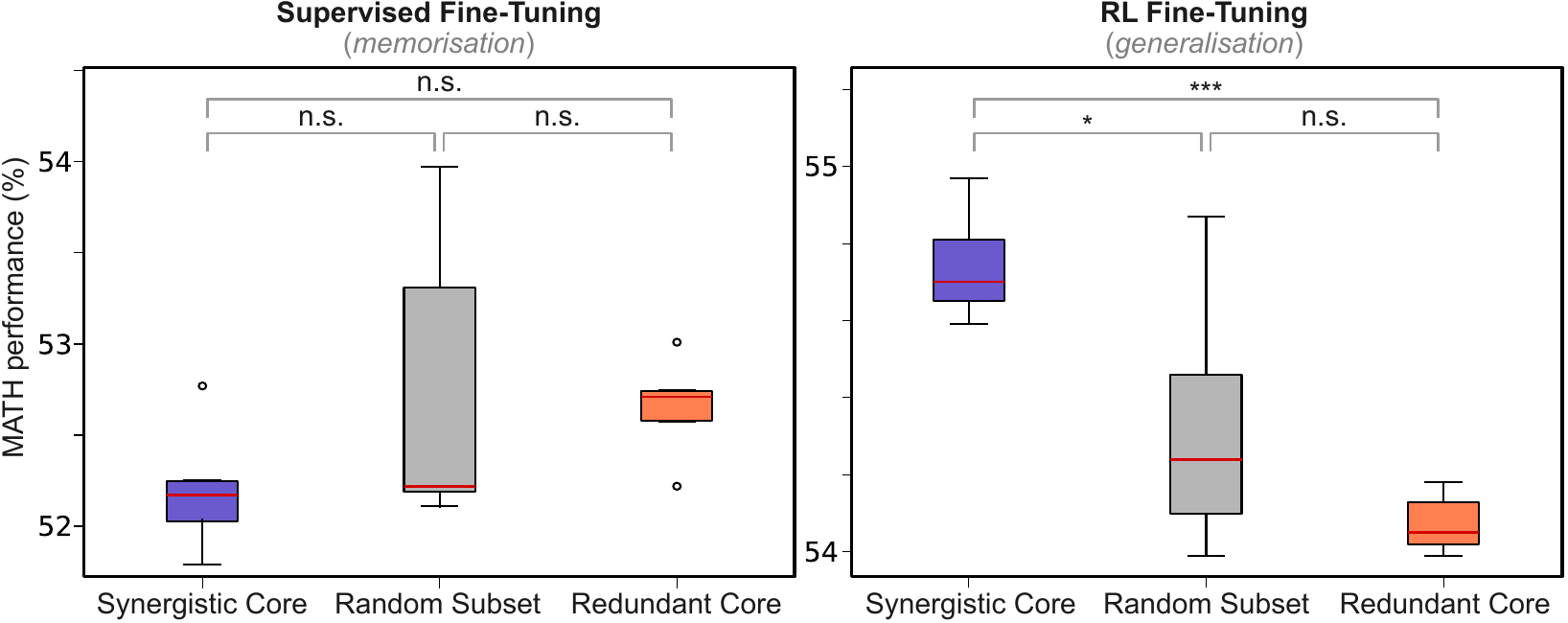}
    \caption{\small
        \textbf{Fine-tuning the synergistic core yields better performance using reinforcement learning (but not supervised) fine-tuning.} Comparison of MATH benchmark accuracy achieved by Qwen2.5-Math-1.5B after Supervised fine-tuning on OpenMathInstruct-2 (left) or Reinforcement Learning fine-tuning on the MATH training set (right), using three different training approaches: synergistic core, redundant core, and random subset training. Each training method is independently run five times, and the best-performing checkpoint from the last five evaluations (at steps 2500, 3000, 3500, 4000, 4500, and 5000) is selected for RL (for SFT, the final checkpoint coincided with the best).
        Under reinforcement learning (right), synergistic core training shows large improvements over random subset training (Hedges' $g\!\approx\!1.4$) and redundant core training (Hedges' $g\!\approx\!5.0$), while no significant differences are observed under supervised fine-tuning (left). Asterisks indicate statistical significance: *$p < 0.05$, **$p < 0.01$, and ***$p < 0.001$ (n.s. represents ``not significant'', $p > 0.05$).
    }
    \label{fig:rlt}
\end{figure}

\section*{Discussion}
\label{sec:discussion}

\subsection*{Convergent Principles of Biological and Artificial Intelligence }

Viewing the brain as an information-processing system has been a fruitful approach in neuroscience~\cite{quian2009extracting, amico2021toward, panzeri2022structures, luppi2024information}, yet the traditional information-theoretic methods that have been widely used to study neural coding often fail to distinguish genuine information integration from mere redundancy. The PID framework~\cite{pid2010beer} overcomes this challenge by disentangling synergy and redundancy, providing a more fine-grained picture of information flow in neural systems~\cite{luppi2024information}. This substrate-independent framework enables a rigorous comparison between biological and artificial intelligence, allowing us to identify fundamental organizational principles that transcend specific physical architectures -- an important and rapidly growing line of work~\cite{Tax2017,ehrlich2023complexity,makkeh2025general}.

Our results add to this growing research body by revealing striking parallels between LLMs -- today's most prominent AI systems -- and recent findings about the informational architecture of the human brain. First, we find that in both brains and LLMs, synergy and redundancy are not in balance, but rather some units are synergy-dominated, and others redundancy-dominated \cite{luppi2022synergistic, varley2023}. Nor is this preference organised randomly: rather, redundancy predominates towards the early layers and sensory cortices, as well as towards the late layers and motor cortices (i.e., input and output systems) whereas synergy predominates in the middle between the two extremes (middle layers and association cortices, respectively). Second, high-synergy units are those most closely linked with performance in LLMs, and those exhibiting the greatest evolutionary difference between humans and other, less cognitively sophisticated primates \cite{luppi2022synergistic}. Third, we find that networks of redundancy are modularly organised, whereas both biological and artificial networks of synergy give preference to efficiency \cite{luppi2022synergistic}. Finally, we find that in LLMs, the high-synergy units are those with greatest relevance for performance when lesioned. The same is true in the human brain: high-synergy regions are the most vulnerable to both pathological (lesion-induced) and pharmacological forms of loss of consciousness \cite{luppi2024workspace}. 

These convergent findings generate specific, testable predictions for neuroscience. Our results in Figure~\figsubref{fig:graph}{a} suggest that the synergistic core in the human brain is an acquired, not innate, pattern. Accordingly, we would predict the emergence of synergistic core during development -- which is already supported by preliminary evidence~\cite{varley2025emergence}. 
At the same time, our results show that the synergistic core has a key role in generalisation. Extrapolating this finding to the brain, brain stimulation to the synergistic core should impair a subject's ability to transfer a learned rule to a new context, without impairing their ability to execute the learned task itself.
Finally, since synergy is more fragile and susceptible to perturbation~\cite{mediano2022integrated}, we expect the synergistic core regions to be particularly vulnerable to neuropathology, which again fits with known clinical reports of frontotemporal dementia~\cite{bang2015frontotemporal} and significant synaptic loss in Alzheimer's Disease~\cite{dekosky1990synapse}

\subsection*{The Anatomy of Transformer Models and the Challenge of Generalization}

Our identification of a synergistic core in the middle layers of LLMs aligns with the emerging consensus on the functional anatomy of the Transformer architecture. It is widely understood that early layers are responsible for detokenization and constructing local feature representations, while late layers handle tokenization and output prediction~\cite{elhage2021mathematical}. However, the specific computational role of the intermediate layers has been a subject of recent debate, which our findings help resolve.

A growing body of literature has suggested that middle layers in deep transformers are largely redundant and can be pruned without significant performance loss. For example, recent work on ``ShortGPT'' argues that layers in LLMs are more redundant than expected~\cite{men2025shortgpt}, while others have demonstrated that models can learn to skip middle layers entirely~\cite{lawson2025learning} or that these layers share a common representation space that allows for their removal~\cite{sun2025transformer}. Some studies even suggest that early layers contribute more to model function than intermediate ones~\cite{singhal2025impact}.

In contrast, an opposing stream of research highlights the critical importance of these layers for higher-order capabilities. Intermediate layers have been found to encode richer representations than their peripheral counterparts~\cite{skean2025layer}, and resource allocation analysis suggests that concentrating parameters in the middle is optimal for performance~\cite{ikeda2025layerwise}. Crucially, mechanistic interpretability studies have extracted causally relevant features primarily from these depths~\cite{templeton2024scaling}, and recent findings indicate that intermediate layers are specifically responsible for out-of-distribution (OOD) generalization~\cite{uselis2025intermediate}.

Our information-theoretic analysis bridges these perspectives. We show that while middle layers may lack the information required for immediate input/output mapping (making them appear dispensable for simple tasks), they are the primary seat of complex information integration. This explains why they are critical for generalization: synergy is required to synthesize disparate information sources into novel representations. This interpretation is strongly supported by our experimental results, where ablating synergistic components caused disproportionate damage to mathematical reasoning (a proxy for generalization), and targeted reinforcement learning of the synergistic core significantly improved performance where supervised fine-tuning failed. Thus, the ``redundancy'' observed in previous pruning studies may be an artifact of evaluating models on tasks that do not require the high-level information integration provided by the synergistic core.

\subsection*{A New Frontier for Model Interpretability}

The emerging field of mechanistic interpretability aims to reverse-engineer the algorithms and mechanisms that deep learning models develop spontaneously. Pioneering work in this domain has focused on identifying specific sub-graphs or ``circuits'' responsible for distinct behaviours~\cite{cammarata2020thread}, leading to the discovery of induction heads for in-context learning~\cite{olsson2022context}, the study of superposition~\cite{elhage2022superposition}, and the use of sparse autoencoders to disentangle polysemantic features~\cite{bricken2023monosemanticity}. These ``bottom-up'' approaches have been remarkably successful at isolating specific functional units, yet they often face challenges in explaining how these units cooperate to produce coherent global behaviour -- particularly when computations are distributed and non-linear.

Our work complements these fine-grained analyses with a ``top-down'' information-theoretic perspective. By quantifying the statistical structure of interactions rather than the content of individual features, we identify robust organizational principles, such as the synergistic core, that link model components to high-level capabilities like generalization. This aligns with prior work showing that learned representations transition from redundancy to synergy during training~\cite{Tax2017,Shujian2021cnns}, suggesting a fundamental shift from independent predictive features to cooperative representations.

Crucially, the framework of information decomposition opens a new avenue for bridging these global and local perspectives: the use of pointwise (or local) information measures~\cite{lizier2012local}. While this study focused on average synergy and redundancy to characterize the macroscopic architecture of LLMs, the same mathematical formalism can be applied at the level of individual tokens and timesteps. This offers a powerful new tool for mechanistic interpretability, enabling the explicit detection of high-order interactions where a model's output depends on the \textit{joint} state of multiple attention heads or neurons, rather than their additive contributions. By using local PID measures to identify exactly when and where synergistic integration occurs for specific prompts, future work can guide circuit discovery towards the complex, multi-component computations that likely underlie the most sophisticated reasoning capabilities of large language models.

\subsection*{Limitations and Future Work}

While our results provide compelling evidence for a synergistic core in LLMs, several limitations of our current approach warrant discussion. First, our analysis focused primarily on attention heads and Mixture-of-Experts (MoE) modules as the fundamental units of information processing. However, Multilayer Perceptrons (MLPs) constitute a significant portion of transformer parameters and are hypothesized to store factual knowledge. Future work should apply information decomposition techniques to MLP layers -- potentially extending recent methodologies~\cite{proca2024,ehrlich2023complexity} -- which may reveal additional or complementary synergistic structures.

Second, our reliance on the L2 norm of attention outputs as a proxy for ``activation'' is a simplification. While effective for identifying macroscopic patterns, this scalar metric cannot capture the full multi-dimensional complexity of the information encoded within attention heads. More refined metrics that account for the vector-space geometry of representations could yield deeper insights into the specific content being integrated.

Third, due to computational constraints, our causal learning experiments were restricted to fine-tuning (RLFT) rather than full-scale pretraining. While we hypothesize that the synergistic core begins to form during pretraining -- a view supported by our observational analysis of Pythia-1B -- validating whether targeted synergistic updates can accelerate initial model convergence remains an open question for future research. Finally, while our custom cognitive task categories provided a controlled environment for testing, expanding this analysis to standard, large-scale benchmarks would allow for more rigorous performance assessments and clearer comparisons with the broader literature.

\section*{Conclusion}\label{sec:conclusion}

In this work, we introduce a neuroscience-inspired, information-theoretic approach to LLM interpretability, using the framework of Integrated Information Decomposition \cite{phid2021}. By treating attention heads as fundamental information-processing units, we show that LLMs exhibit a \emph{synergistic core} in their middle layers and a predominantly \emph{redundant periphery} in their early and late layers. This architecture closely mirrors recent neuroscience findings in the human brain \cite{luppi2022synergistic} which reports higher synergy in brain regions further away from the sensory periphery. Furthermore, LLMs' synergistic core emerges dynamically through training and shares large-scale topological properties with human brains -- overall suggesting a convergent information-processing hierarchy in both biological and artificial intelligences.

Importantly, our causal interventions -- through both ablation and RL-based fine-tuning -- confirm that this synergistic core is a primary driver of model performance and generalisation. Thus, the synergistic core is not merely a byproduct of the model's structure, but it can be causally manipulated to either impair or enhance the model's performance.
These results further validate synergy as a key metric for understanding how large models generalize and leverage information for downstream tasks. 

Ultimately, this framework offers a powerful top-down alternative to fine-grained approaches in mechanistic interpretability. By identifying robust organizational principles that link model components to high-level behaviour, we provide a foundation for a novel line of research on information-theoretic LLM interpretability. Bridging neuroscience and deep learning not only enriches our understanding of large-scale models but also offers a rigorous lens through which to explore the broader ``space of possible minds''~\cite{sloman1984structure} and the universal principles of intelligence.

\section*{Methods}\label{sec:methods}

This work builds upon foundational concepts from Information Theory (IT) and its recent extensions, namely Partial Information Decomposition (PID) and Integrated Information Decomposition ($\Phi$ID).

\subsection*{Partial Information Decomposition}

In classical information theory \cite{shannon_1948}, the mutual information between random variables $X$ and $Y$ quantifies the amount of information shared between them and serves as a generalised measure of dependency. It is defined as 
\begin{equation*}
    I(X;Y) = \mathbb{E}_{X,Y} \left[\log \frac{p(x,y)}{p(x)p(y)}\right] = H(Y) - H(Y|X).
\end{equation*}

Partial Information Decomposition (PID) is an extension of classical IT introduced by Beer et al. in \cite{pid2010beer}. Classical IT lacks a notion of \textit{sameness} of information, which PID aims to address.

To this end, PID considers the setting where we have two source random variables $X_1$ and $X_2$ that contain information about a target random variable $Y$, such that $I(Y;X_1, X_2)>0$. PID seeks to disentangle how each of the source random variables contributes to the target random variable.

Classical IT does not provide the conceptual tools to achieve this, as $I(Y;X_1, X_2) \neq I(Y;X_1) + I(Y;X_2)$. On the one hand, if $X_1=X_2$, adding their mutual information would count the information twice, which is referred to as redundant information. On the other hand, there may be information provided only when $X_1$ and $X_2$ are considered jointly and not by either in isolation, which is referred to as synergistic information.

To address these shortcomings, PID introduces three distinct notions of information, illustrated in Figure \ref{fig:conceptual}:
\begin{itemize}
    \item \textbf{Redundant Information:} $I_{red}(Y;X_1, X_2)$. Information about the target that is present repeatedly in both source variables.
    \item \textbf{Unique Information:} $I_{unq}(Y;X_1 \setminus X_2)$ and $I_{unq}(Y;X_2 \setminus X_1)$. Information about the target that is specific to one source and absent in the other.
    \item \textbf{Synergistic Information:} $I_{syn}(Y;X_1, X_2)$. Information about the target that emerges only when both source variables are considered together, not discernible when analyzing the sources individually.
\end{itemize}

These quantities, collectively known as PID atoms, provide a comprehensive decomposition of the mutual information $I(Y; X_1, X_2)$. Specifically, the decomposition satisfies the following equations:
\begin{align*}
     I(Y; X_1, X_2) &= I_{red}(Y;X_1, X_2) + I_{unq}(Y;X_1\setminus X_2)+ I_{unq}(Y;X_2\setminus X_1) + I_{syn}(Y;X_1, X_2), \\
     I(Y; X_1) &= I_{unq}(Y;X_1\setminus X_2)+ I_{red}(Y;X_1, X_2),  \\ 
     I(Y; X_2) &= I_{unq}(Y;X_2\setminus X_1)+ I_{red}(Y;X_1, X_2). 
\end{align*}

However, classical IT quantities alone are insufficient to compute these PID atoms, as they lead to an underdetermined system of equations. To resolve this, it is necessary to define a specific method for computing one of the PID atoms. Following Beer et al. \cite{pid2010beer}, we compute the redundant information as:
\begin{equation*}
    I_{red}(Y; X_1, X_2) = I_{min}(Y; X_1, X_2) = \sum_{y\in \mathcal{Y}} p(y) \min\{ I(Y=y; X_1), I(Y=y; X_2)\}.
\end{equation*}

\subsection*{Integrated Information Decomposition}

While PID addresses the decomposition of information provided by several source variables about a single target variable, an ideal extension would allow this decomposition to span multiple targets. This is precisely the motivation for Integrated Information Decomposition ($\Phi$ID), introduced by Mediano et al. in \cite{phid2021}.

$\Phi$ID enables decomposing the Time-Delayed Mutual Information (TDMI):
\begin{equation*}
    \text{TDMI} = I(X_t^1, X_t^2; X_{t+1}^1, X_{t+1}^2),
\end{equation*}
where the superscripts $1$ and $2$ denote two distinct components or modules of a complex system. Decomposing this quantity enables a fine-grained analysis of the \textit{information dynamics} between parts of the system over time.

$\Phi$ID interprets these dynamics as temporal rearrangements of PID atoms. For instance, information encoded synergistically in the sources can become uniquely encoded in one of the targets. We denote this as $Syn\rightarrow Unq^1$ and corresponds to the notion of downward causation. Similarly, information uniquely encoded by a single source may become redundantly encoded across both target variables. We denote this information as $Red\rightarrow Unq^2$ and corresponds to the notion of copying. More generally, $\Phi$ID introduces $4^2=16$ temporal combinations of PID atoms which form the lattice of information atoms.

\subsection*{Estimating Synergy and Redundancy in Large Language Models}

To investigate the informational architecture of LLMs, we adopt several neuroscientifically inspired methodological choices, guided by the approach of Luppi et al. \cite{luppi2022synergistic}.

The first step in our information-theoretic analysis involves selecting appropriate units, or nodes, for study. Attention heads and experts are chosen as the primary information-processing units due to their functional analogy to distinct brain regions. We utilize the $\Phi$ID framework to characterize the informational properties of these nodes. However, unlike the human brain, which possesses a natural, intrinsic notion of time, transformer architectures permit multiple temporal interpretations. We define each autoregressive generation step explicitly as a discrete timestep to capture the temporal evolution of information processing.

To perform the $\Phi$ID analysis, we require a well-defined notion of the \textit{activation} of an attention head at each autoregressive timestep. For a given attention head $h_i$, we define its activation as the L2 norm of its attention output vector
\begin{equation*}
    a(h_i, t) := \left\| \text{softmax}\left(\frac{q_i^t (K_i^t)^\top}{\sqrt{d_k}}\right) V_i^t \right\|_2,
\end{equation*}
where $q_i^t$ is the query vector for head $h_i$ at time $t$, $K_i^t, V_i^t$ are the key and value matrices for head $h_i$ at time $t$, and $d_k$ is the head dimension. For experts, the activation is analogously defined as the L2 norm of the expert output vector.

For the synergy and redundancy analysis, we adopt the approach of Luppi et al. \cite{luppi2022synergistic}, using the temporally persistent synergy ($Syn\rightarrow Syn$) and temporally persistent redundancy ($Red\rightarrow Red$) atoms in $\Phi$ID as our measures of synergy and redundancy. These metrics capture persistent interactions over time, enabling us to identify where the most synergistic and redundant interactions occur within the model.

We conduct our analysis using several open-source LLM families, including Gemma 3 \cite{google2024gemma}, Llama 3 \cite{dubey2024llama}, Qwen 3 \cite{yang2025qwen3}, and DeepSeek V2 Lite Chat \cite{liu2024deepseek}. For the DeepSeek model, we utilize its experts as the nodes for our information-theoretic analysis, whereas we employ attention heads for the remaining models.

\begin{table}
    \centering
    \begin{tabular}{>{\centering\arraybackslash}p{5cm} >{\centering\arraybackslash}p{10cm}}
        \toprule
        \textbf{Cognitive Task Category} & \textbf{Example Prompt} \\
        \midrule
        Syntax and Grammar Correction & Correct the error: He go to school every day. \\
        \addlinespace
        Part of Speech Tagging & Identify the parts of speech in the sentence: Quickly, the agile cat climbed the tall tree. \\
        \addlinespace
        Basic Numerical Reasoning & If you have 15 apples and you give away 5, how many do you have left? \\
        \addlinespace
        Basic Common Sense Reasoning & If it starts raining while the sun is shining, what weather phenomenon might you expect to see? \\
        \addlinespace
        Abstract Reasoning and Creative Thinking & Imagine a future where humans have evolved to live underwater. Describe the adaptations they might develop. \\
        \addlinespace
        Emotional Intelligence and Social Cognition & Write a dialogue between two characters where one comforts the other after a loss, demonstrating empathy. \\
        \bottomrule
    \end{tabular}
    \caption{Cognitive task categories and example prompts, sorted by increasing cognitive complexity. A full list of the prompts used for our analysis is found in Appendix~\ref{appendix:prompts}.}
    \label{tab:cognitive_task_examples}
\end{table}

\subsection*{Computing Behaviour Divergence}\label{sec:behaviour-divergence-methods}

Let $\mathcal V$ be the model's vocabulary and let $q\in \mathcal Q$ be a prompt from our set of prompts $\mathcal Q$. Let $ \mathbf{x}_{<t}^{(na)} $ denote the sequence of tokens generated up to time step $ t-1 $ by the non-ablated model when answering a prompt $q$. At each time step $ t $, the non-ablated and ablated models induce probability distributions over the next token $ x_t $, $ \mathbf{p}_t^{(na)}(x_t | \mathbf{x}_{<t}^{(na)})$ and $\mathbf{p}_t^{(a)}(x_t | \mathbf{x}_{<t}^{(na)})$. The KL divergence between these distributions at time step $ t $ for a given prompt $q\in \mathcal Q$ is computed as:
\begin{equation*}
D_{\text{KL}}^q(\mathbf{p}_t^{(na)} \| \mathbf{p}_t^{(a)}) = \sum_{x_t \in \mathcal V} \mathbf{p}_t^{(na)}(x_t | \mathbf{x}_{<t}^{(na)}) \log \frac{\mathbf{p}_t^{(na)}(x_t | \mathbf{x}_{<t}^{(na)})}{\mathbf{p}_t^{(a)}(x_t | \mathbf{x}_{<t}^{(na)})}.
\end{equation*}

We use this divergence as a task-agnostic proxy for behavioural impact, which is particularly appropriate for open-ended prompts where correctness cannot be unambiguously defined.

This approach explicitly conditions the ablated model on the sequence of tokens generated by the non-ablated model, ensuring that the comparison between the two models focuses on the differences in their probability distributions rather than on compounding effects due to differing input sequences. 

To compute the overall behaviour divergence across a prompt $q\in \mathcal Q$ response sequence of $ T $ tokens (excluding the prompt), we average the KL divergence over all tokens:
\begin{equation*}
\text{Behaviour divergence}(q) = \frac{1}{T} \sum_{t=1}^{T} D^q_{\text{KL}}(\mathbf{p}_t^{(na)} \| \mathbf{p}_t^{(a)}).
\end{equation*}

Higher behaviour divergence indicates a greater causal impact of the ablation on the model’s output distribution.

Unless otherwise stated, we compute this measure while iteratively ablating attention heads either in decreasing order of synergy–redundancy rank or according to a random ordering, enabling a comparison between targeted and untargeted perturbations.

\subsection*{Reinforcement Learning Fine-Tuning Procedure}\label{sec:rl-procedure}

Training consists of 5{,}000 GRPO steps with 8 sampled trajectories per step, a maximum token length of $1024$, and an effective batch size of 16. For each of the three training conditions (synergistic core, redundant core, random subset), we perform five independent runs. Model performance is evaluated on the MATH benchmark, and for each run we report the best-performing checkpoint among evaluations at steps 2500, 3000, 3500, 4000, 4500, and 5000.

\newpage
\bibliographystyle{IEEEtran}
\bibliography{bibliography}
\newpage

\appendix

\section{Prompts used for the Different Cognitive Task Categories} \label{appendix:prompts}

\begin{table}[!ht]
    \begin{adjustwidth}{-0.25cm}{0.25cm}  
    \centering
    \begin{tabular}{>{\centering\arraybackslash}p{2.5cm} >{\centering\arraybackslash}p{13.5cm}}
        \toprule
        \textbf{Cognitive Task Category} & \textbf{Prompts} \\
        \midrule
        \multirow{10}{=}{\textbf{Syntax and Grammar Correction}} & Correct the error: He go to school every day. \\
                                      & Correct the error: She have two cats and a dogs. \\
                                      & Correct the error: I eats breakfast at 8:00 in the morning. \\
                                      & Correct the error: Every students in the classroom has their own laptop. \\
                                      & Correct the error: She don't like going to the park on weekends. \\
                                      & Correct the error: We was happy to see the rainbow after the storm. \\
                                      & Correct the error: There is many reasons to celebrate today. \\
                                      & Correct the error: Him and I went to the market yesterday. \\
                                      & Correct the error: The books is on the table. \\
                                      & Correct the error: They walks to school together every morning. \\
        \midrule
        \multirow{10}{=}{\textbf{Part of Speech Tagging}} & Identify the parts of speech in the sentence: Quickly, the agile cat climbed the tall tree. \\
                               & Identify the parts of speech in the sentence: She whispered a secret to her friend during the boring lecture. \\
                               & Identify the parts of speech in the sentence: The sun sets in the west. \\
                               & Identify the parts of speech in the sentence: Can you believe this amazing view? \\
                               & Identify the parts of speech in the sentence: He quickly finished his homework. \\
                               & Identify the parts of speech in the sentence: The beautifully decorated cake was a sight to behold. \\
                               & Identify the parts of speech in the sentence: They will travel to Japan next month. \\
                               & Identify the parts of speech in the sentence: My favorite book was lost. \\
                               & Identify the parts of speech in the sentence: The loud music could be heard from miles away. \\
                               & Identify the parts of speech in the sentence: She sold all of her paintings at the art show. \\
        \midrule
        \multirow{10}{=}{\textbf{Basic Numerical Reasoning}} & If you have 15 apples and you give away 5, how many do you have left? \\
                                  & A rectangle's length is twice its width. If the rectangle's perimeter is 36 meters, what are its length and width? \\
                                  & You read 45 pages of a book each day. How many pages will you have read after 7 days? \\
                                  & If a train travels 60 miles in 1 hour, how far will it travel in 3 hours? \\
                                  & There are 8 slices in a pizza. If you eat 2 slices, what fraction of the pizza is left? \\
                                  & If one pencil costs 50 cents, how much do 12 pencils cost? \\
                                  & You have a 2-liter bottle of soda. If you pour out 500 milliliters, how much soda is left? \\
                                  & A marathon is 42 kilometers long. If you have run 10 kilometers, how much further do you have to run? \\
                                  & If you divide 24 by 3, then multiply by 2, what is the result? \\
                                  & A car travels 150 miles on 10 gallons of gas. How many miles per gallon does the car get? \\
        \bottomrule
        \bottomrule
    \end{tabular}
    \caption{Expanded list of prompts used for the different cognitive task categories, Part 1.}
    \label{tab:full_cognitive_task_examples_part1}
\end{adjustwidth}
\end{table}

\begin{table}[ht]
    \centering
    \begin{tabular}{>{\centering\arraybackslash}p{2.5cm} >{\centering\arraybackslash}p{13cm}}
        \toprule
        \textbf{Cognitive Task Category} & \textbf{Prompts} \\
        \midrule

        \addlinespace
        \multirow{10}{=}{\textbf{Basic Common Sense Reasoning}} & If it starts raining while the sun is shining, what weather phenomenon might you expect to see? \\
                                     & Why do people wear sunglasses? \\
                                     & What might you use to write on a chalkboard? \\
                                     & Why would you put a letter in an envelope? \\
                                     & If you're cold, what might you do to get warm? \\
                                     & What is the purpose of a refrigerator? \\
                                     & Why might someone plant a tree? \\
                                     & What happens to ice when it's left out in the sun? \\
                                     & Why do people shake hands when they meet? \\
                                     & What can you use to measure the length of a desk? \\

        \midrule
        \multirow{10}{=}{\textbf{Abstract Reasoning and Creative Thinking}} & Imagine a future where humans have evolved to live underwater. Describe the adaptations they might develop. \\
                                                  & Invent a sport that could be played on Mars considering its lower gravity compared to Earth. Describe the rules. \\
                                                  & Describe a world where water is scarce, and every drop counts. \\
                                                  & Write a story about a child who discovers they can speak to animals. \\
                                                  & Imagine a city that floats in the sky. What does it look like, and how do people live? \\
                                                  & Create a dialogue between a human and an alien meeting for the first time. \\
                                                  & Design a vehicle that can travel on land, water, and air. Describe its features. \\
                                                  & Imagine a new holiday and explain how people celebrate it. \\
                                                  & Write a poem about a journey through a desert. \\
                                                  & Describe a device that allows you to experience other people's dreams. \\
        \midrule
        \multirow{10}{=}{\textbf{Emotional Intelligence and Social Cognition}} & Write a dialogue between two characters where one comforts the other after a loss, demonstrating empathy. \\
                                                    & Describe a situation where someone misinterprets a friend's actions as hostile, and how they resolve the misunderstanding. \\
                                                    & Compose a letter from a character apologising for a mistake they made. \\
                                                    & Describe a scene where a character realizes they are in love. \\
                                                    & Write a conversation between two old friends who haven't seen each other in years. \\
                                                    & Imagine a character facing a moral dilemma. What do they choose and why? \\
                                                    & Describe a character who is trying to make amends for past actions. \\
                                                    & Write about a character who overcomes a fear with the help of a friend. \\
                                                    & Create a story about a misunderstanding between characters from different cultures. \\
                                                    & Imagine a scenario where a character has to forgive someone who wronged them. \\
        \bottomrule
        \bottomrule
    \end{tabular}
    \caption{Expanded list of prompts used for the different cognitive task categories, Part 2.}
    \label{tab:full_cognitive_task_examples_part2}
\end{table}

\clearpage

\end{document}